\documentclass{llncs}
\usepackage{makeidx}  % allows for indexgeneration
\usepackage{latexsym}
\usepackage{graphicx}
\usepackage{algorithm2e}
\usepackage{subfig}
\usepackage{verbatim}

\newcommand{\keywords}[1]{\par\addvspace\baselineskip
\noindent\keywordname\enspace\ignorespaces#1}
\newcommand{\be}{\begin{equation}}
\newcommand{\ee}{\end{equation}}
\newcommand{\bea}{\begin{eqnarray}}
\newcommand{\eea}{\end{eqnarray}}

\newcommand{\bee}{\begin{eqnarray*}}
\newcommand{\eee}{\end{eqnarray*}}
\begin{document}
\title{\bf A Review of Object Detection Models based on Convolutional Neural Network}
\author{F. Sultana \inst{1} \and A. Sufian \inst{1, *}  \and P. Dutta \inst{2} \\}
\institute{Department of Comp. Sc., Uiversity of Gour Banga, WB, India.\\
			 \and 
Department of Comp. \& Sys. Sc., Visva-Bharati University, WB, India} 					
\maketitle
\begin{center}
\footnotesize{
	\(^*\)Corresponding Author (e-mail: sufian.csa@gmail.com)} 
\end{center}
\begin{abstract}
 Convolutional Neural Network (CNN) has become the state-of-the-art for object detection in image task. In this chapter, we have explained different state-of-the-art CNN based object detection models. We have made this review with categorization those detection models according to two different approaches: two-stage approach and one-stage approach. Through this chapter, it has shown advancements in object detection models from R-CNN to latest RefineDet. It has also discussed the model description and training details of each model. Here, we have also drawn a comparison among those models. 
 
\keywords{Convolutional Neural Network, Deep Learning, Mask R-CNN, Object Detection, R-CNN, RetinaNet, Review, YOLO}
\end{abstract}
{\section{Introduction}

In computer vision, the problem of estimating the class and location of objects contained within an image is known as object detection problem. Unlike classification, every instances of objects are detected in the task of object detection. So object detection is basically instance-wise vision task. Before the popularity of deep learning in computer vision, object detection was done by using hand crafted machine learning features such as shift invariant feature transform (SHIFT) \cite{Lowe2004}, histogram of oriented gradients (HOG) \cite{Dalal2005} etc. At that era, the best object detection models gain around 35\% mean average precision (mAP) on PASCAL VOC Challenge dataset \cite{Everingham2010}. But during 2010-2012, the progress in object detection has been quite motionless. The best performing models gained their result building complex ensemble systems and applying few modification in successful models. Also the representation of SHIFT and HOG could only associate with the complex cells in V1 of Primary Visual Cortex. But both the structure of visual cortex of animals and the object recognition process in computer vision found the fact that there might be hierarchical, multi-level processes for computing features that may contain more information for visual recognition. Inspired by this idea Kunihiko Fukushima built ``neocognitron" \cite{Fukushima1980} which is a hierarchical, shift-invariant multi-layered neural network for visual pattern recognition. In 1990, LeCun et al. extended the theory of neocognitron and introduced the first practical image classification model based on convolutional neural network called  ``LeNet-5" \cite{LeCun1989}\cite{LeCun1998}. This model was trained using supervised learning through stochastic gradient descent (SGD) \cite{Bottou2010} via backpropagation (BP) algorithm \cite{LeCun1988}. After that, the progress of CNN was stagnant for several years. In 2012, the resurgence of ``AlexNet" \cite{Krizhevsky2012} inspired researchers of different field of computer vision such as image classification, localization, object detection, segmentation etc. and we have got many different state-of-the-art classification models \cite{Sultana2018} and object detection models like R-CNN \cite{Girshick2014}, SPP-net \cite{He2014}, Fast R-CNN \cite{Girshick2015}, YOLO \cite{Redmon2015}, RetinaNet \cite{Lin2017} etc. in successive years.% In this paper we have presented brief overview of uses of CNN in object detection problem of computer vision. 

In this chapter, we have wrote a review of different state-of-the-art Convolutional Neural Network(CNN)\cite{CNN_Chapter19} based object detection models. We have described network architecture along with training details of various models in section 2 with some contextual subsections. In section 3,  we have compared the performance of those object detection models on different datasets. Finally, we have concluded our chapter in section 4.

\section{Different Object Detection Models}

After 2012, different researchers applied different strategies such as object proposals algorithm, pooling methods, novel loss functions, bounding box estimation methods etc along with convolutional neural network for betterment of the task of object detection. The CNN based object detection models can be categorized into two different categories: (i) two stage approach and (ii) one stage approach. Models of these two categories has discussed in the following two subsections.  
\subsection{Two Stage approach:}
In the two stage object detection, first stage generates region or object proposals and in second stage those proposals are classified and detected using bounding boxes. The state-of-the-art models falls in this category has explained one-by-one:
\begin{figure}[htb]
\vspace{-.15in}
	\centering
	\includegraphics[scale=0.5]{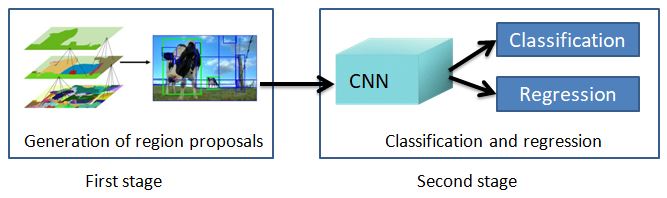}
	\caption{Two stage approach}
	\label{ftwo_stage}
\end{figure}
\vspace{-.15in}
 %<<==========================R-CNN===================================>>
\vspace{-.10in}
\subsubsection{R-CNN:}		\vspace{-.10in}
Region proposal with convolutional neural network (R-CNN) \cite{Girshick2015} is the first CNN based two stage object detection model. The architecture of R-CNN contains three different blocks as shown in figure \ref{frcnn}. First stage includes the first block where the authors have used selective search \cite{Uijlings2013} algorithm to generate around 2000 class-independent region proposals from each input image. In the second block, following the architecture of AlexNet \cite{Krizhevsky2012} they have used a CNN with five convolutional (Conv) layers and two fully connected (FC) layers to extract fixed length feature vector from each region proposal. As CNN requires fixed sized image as input, the authors have used affine image warping \cite{Wolberg1994} to get fixed sized input image from each region proposals regardless of their size or aspect ratio. Then those warped images are fed into individual CNN to extract fixed length feature vectors from each region proposal. The third block classifies each region proposal with category specific linear support vector machine (SVM) \cite{Hearts1998}. Second and third block together constitute the second stage.

%\begin{figure*}
%\vspace{-.15in}
%\centering
%\subfloat[]{\includegraphics[width=5cm]{image/rcnn3}\label{frcnn}}
%\hspace{2mm}
%\subfloat[]{\includegraphics[width=5cm]{image/sppnet2}\label{fsppnet}}\\
%\caption{ Arcitecture of (a) R-CNN\cite{Girshick2015} and (b)  SPP-net}

%\label{frcnn_sppnet}	
%\end{figure*}

	\begin{figure}[htb]
		\vspace{-.15in}
		\centering
		\includegraphics[scale=0.65]{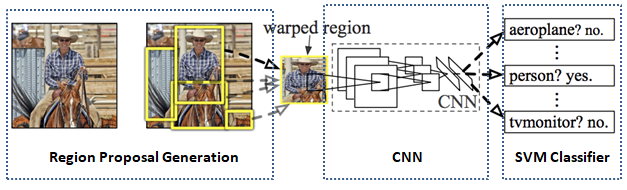}
		\caption{Arcitecture of R-CNN\cite{Girshick2015} }
		\label{frcnn}
		\vspace{-.10in}
	\end{figure}

\paragraph{\textbf{ Training details:}}
The authors of R-CNN have pre-trained the CNN of their model using ILSVRC-2012 \cite{Deng2009} \cite{Russakovsky2015} dataset. Then they have changed the ImageNet specific 1000-way softmax classifier with a 21-way classifier (for the 20 PASCAL VOC classes and one background class) and trained the CNN parameters using SGD with warped region proposals taken from PASCAL VOC \cite{Everingham2010} images. If the IoU overlap of region proposals with corresponding ground truth box is $\geq$ 0.5, they treat the proposal as positive for that box class and rest as negative proposal. The learning rate of SGD was 0.001. The mini-batch size per SGD iteration was 128 (32 positive foreground and 96 negetive background window). They have optimized per class linear SVM using standard hard negative mining \cite{Bucher2016} to reduce memory consumption.        

The use of CNN for object detection helped R-CNN to gain higher accuracy in detection problem but there is a major downside of this model.  CNNs need fixed sized input image but region proposals generated by R-CNN are arbitrary. To fulfill the CNNs requirement, the scale or aspect ratio or the originality of an image are got compromised due to cropping, warping, predefined scales etc.
%\vspace{-3mm}

%\vspace{-1.2cm}
 %<<==========================SPP-net===================================>>

\subsubsection{SPP-net:}
In a CNN, Conv  layers actually do not need fixed size image as input. But FC  layers require fixed length feature vectors as input. On the other hand, spatial pyramid pooling (SPP) \cite{Grauman2005}\cite{Lazebnik2006} can generate fixed sized output regardless of input size/scal/aspect ratio. So, He et al includes SPP layer in between conv layers and FC layers of R-CNN and introduced SPP-net \cite{He2014} as shown in figure \ref{fsppnet}. SPP layers pool the features and generates fixed length output from variable length region proposals, which are fed into FC layers for further processing. In this way, SPP-net made it possible to train and test the model with image of varying sizes and scales. Thus, it increases scale invariance as well as reduces overfitting. 
\begin{figure}[htb]
	\vspace{-.15in}
	\centering
	\includegraphics[scale=0.7]{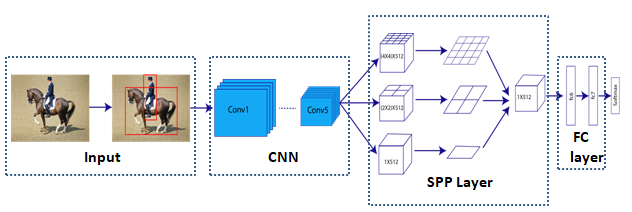}
	\caption{Arcitecture of SPP-net }
	\label{fsppnet}
	\vspace{-.10in}
\end{figure}
\paragraph{\textbf{Training details:}}
The authors have followed \cite{Girshick2014}\cite{VanDeSande2011} to train their SVM classifier. They have also used ground truth box to generate positive proposals. If the IoU of ground truth box and positive window is $\le$ 0.3, then they considered that window as negative sample. If negative sample - negative sample overlap is $\ge$ 0.7, then that sample is removed. They have also applied standard hard negative mining \cite{Felzenswalb2010} to train the SVM.  At test time, the classifier is used to score the candidate windows. Then the authors used non-maximum suppression \cite{Felzenswalb2010} with threshold of 0.3 on the scored windows. The authors have also followed \cite{Girshick2014} to fine tune only the FC layers of their model. The model was trained using SGD. In each mini-batch, the ratio of positive and negative samples was 1:3. They started training 250K mini-batches with the learning rate 0.0001, and then 50K mini-batches with 0.00001. Also following \cite{Girshick2014} they have used bounding box regression on the pooled features from fifth convolutional layer. If the IoU of a window and a ground truth box is $>$ 0.5 then that is used in bounding box regression training.

As SPP-net uses CNN once on the entire image and region proposals are extracted from last conv feature map, the network is much faster than R-CNN.
 %<<==========================Fast R-CNN ==============================>>
\subsubsection{Fast R-CNN:}
R-CNN and SPP-net had some common problems: (i) multi-stage pipeline training (feature extraction, network fine tuning, SVM training, bounding box regression), (ii) expensive training in terms of space and time, and also (iii) slow object detection. Fast R-CNN \cite{Girshick2015}  tries to overcome all of the above limitations. Ross Girshick modified R-CNN and proposed Fast R-CNN which is a single-stage training algorithm that learns to classify region proposals and corrects their spatial locations together. Fast R-CNN can train very deep detection network like VGG-16 \cite{Simonyan2014}, and it becomes  9$\times $ faster than R-CNN and 3$\times$ faster than SPP-net. 

R-CNN used CNN for each generated region proposals for further detection process. But  Fast R-CNN takes entire image and a set of object proposals together as input. From the produced CNN feature map, Region of Interests (RoI) are identified using selective search method. Then the authors have used a RoI pooling layer to reshape the RoIs into a fixed length feature vector. After that, FC layers take those feature vectors as input and passed the output to two sibling output branch. One branch  is for classification and another for bounding box regression. Figure \ref{ffast_rcnn} demonstrates the architecture of Fast R-CNN.
%using a RoI pooling layer we reshape them into a fixed size so that it can be fed into a fully connected layer. From the RoI feature vector, we use a softmax layer to predict the class of the proposed region and also the offset values for the bounding box.
%\begin{figure*}
%		\vspace{-.15in}
%	\centering
%	\subfloat[]{\includegraphics[width=6cm]{image/faster_rcnn}\label{ffaster_rcnn}}\\
%	\caption{ Arcitecture of (a) Fast R-CNN \cite{Girshick2015} and (b)  Faster R-CNN}
%	
%	\label{ffast_faster_rcnn}	
%\end{figure*}

		\begin{figure}[htb]
			\vspace{-.15in}
			\centering
			\includegraphics[scale=0.55]{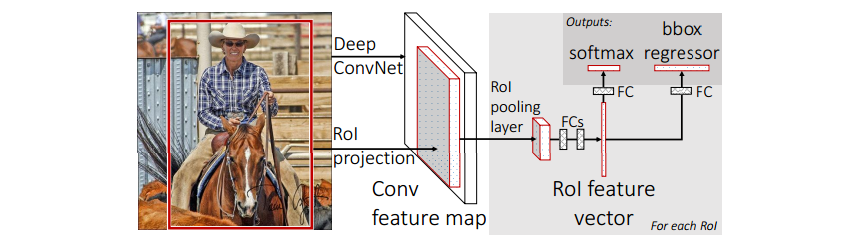}
			\caption{Arcitecture of Fast R-CNN \cite{Girshick2015} }
			\label{ffast_rcnn}
			\vspace{-.10in}
		\end{figure}

\paragraph{\textbf{Training details:}}
The authors have experimented their model using three types of network which are pre-trained on ImageNet \cite{Deng2009} dataset. Three networks are CaffeNet \cite{CaffeNet}, AlexNet \cite{Krizhevsky2012}, and VGG\_CNN\_M\_1024 \cite{Chatfield2014} with another version VGG16 \cite{Simonyan2014}.  The networks have 5 max-pooling layers in total and the number of Conv layers in those network was in between five to thirteen. During initialization of those pre-trained network, the authors have  first replaced the last max-pooling layer with a RoI pooling layer. Then last FC layer and softmax layer of the networks were replaced with two sibling output layers for classification and bounding box regression. The networks are also modified to take two inputs: a whole input image and a list of RoIs present in those images. The authors have trained Fast R-CNN using stochastic gradient descent (SGD) with hierarchically sampled mini-batches. First the authors have chosen $N$ number of image samples randomly. Then from each image,  $R/N$ number of RoIs are sampled for each mini-batch. During fine tuning they have chosen $N=2$ and $R=128$ to construct mini-batch. Initial learning rate for 30K mini-batch iterations was 0.001 and for next 10K mini-batch itertions  was 0.0001. They have also used a momentum of 0.9 and parameter decay of 0.0005. They have used multi-task loss to jointly train their network for classification and bounding box regression in a single stage. They have followed \cite{Girshick2015} and \cite{He2014} to chose the ratio of positive and negative RoIs for training their model. 

Using CNN once for an entire image and single stage training procedure reduces the time complexity of Fast R-CNN in a large scale than previous two R-CNN and SPP-net. Also use of RoI pooling layer and some tricks in training helped their model to achieve higher accuracy.
 %<<==========================Faster R-CNN=============================>>
\subsubsection{Faster R-CNN:} 
R-CNN, SPP-net and Fast R-CNN depend on the region proposal algorithm for object detection. All three models experienced that the computation of region proposals is  time-consuming which affects the overall performance of the network. Ren et al. proposed Faster R-CNN \cite{Ren2015}, where they have replaced previously mentioned region proposal method with region proposal network (RPN). An RPN is a fully convolutional network (FCN) \cite{Long2014} that takes an image of arbitrary size as input and outputs a set of rectangular candidate object proposals. Each object proposal is associated with an objectness scores to detect whether the proposal contains an object or not.
	\begin{figure}[htb]
	\vspace{-.15in}
	\centering
	\includegraphics[scale=0.55]{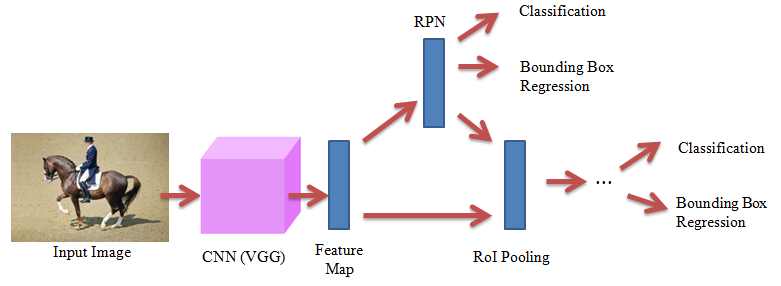}
	\caption{Arcitecture of  Faster R-CNN }
	\label{ffaster_rcnn}
	\vspace{-.10in}
\end{figure}

Similar to Fast R-CNN, the entire image is provided as an input to the Conv layers of Faster R-CNN to produce convolutional feature map. Then instead of using selective search algorithm on the feature map to identify the region proposals, a RPN is used to predict the region proposals. In RPN, the region proposals detected per sliding window locations are called anchors. An relevant anchor box is selected by applying a threshold value over the ``objectness'' score. Selected anchor boxes and the feature maps computed by the initial CNN model together are fed to RoI pooling layer for reshaping and the output of RoI pooling layer fed into FC layers for final classification and bounding box regression. 

\paragraph{\textbf{Training details:}}
The RPN is basically  a fully convolutional neural network \cite{Shelhamer2017} pre-trained over the ImageNet dataset and it is fine-tuned on the PASCAL VOC dataset. Generated region proposals from RPN with anchor boxes are used to train the Fast R-CNN (later part of Faster R-CNN after RPN). To train RPN, each anchor is assigned to a binary class label. Positive label is assigned to two types of anchors: `` (i) the anchor/anchors with the highest Intersection-over-Union (IoU) overlap with a ground-truth box, or (ii) an anchor that has an IoU overlap $>$ 0.7 with any ground-truth box''. A negative label is assigned to an anchor if its IoU value is $<$ 0.3 for all ground-truth boxes. The authors have trained both RPN and Fast R-CNN independently. They simply followed the multi-task loss of Fast R-CNN to train their network. RPN is end-to-end trainable with back propagation and SGD. The authors have also followed the `image centric' sampling strategy of Fast R-CNN. A mini-batch of SGD is constructed with a number of positive and negative anchor boxes predicted from an input image. In each mini-batch, the ratio of positive and negetive anchor could have a ratio upto 1:1. They have used a initial learning rate of 0.001 for 60K min-batches and 0.0001 for the next 20K mini-batches. The authors have used 0.9 as momentum and 0.0005 as weight decay.

Previously mentioned object detector models considered single scale feature map for object detection.  In this way those models lack exact location of object instances as the last layer feature map of CNN is scale invariant.   
%%=====================FPN=======================%%
\subsubsection{FPN:}
Lin et al used the multilevel feature map of CNN in a different way to construct Feature Pyramid Network(FPN) \cite{Lin2016_FPN} for better object detection than state-of-the-art models. 
    \begin{figure}[htb]
    	\centering
    	\includegraphics[scale=0.4]{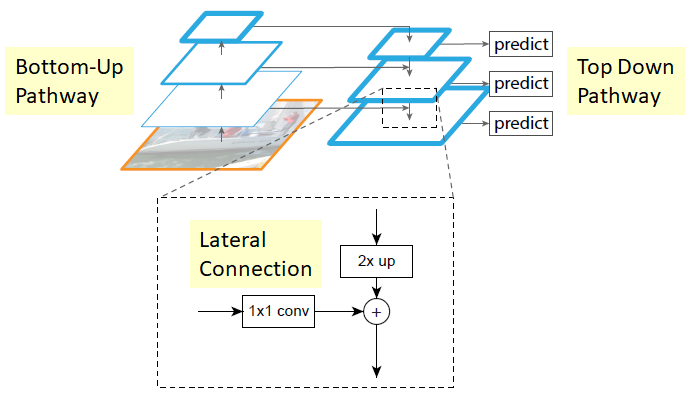}
    	\caption{Bottom-up and top-down architecture of FPN with a building block \cite{Lin2016_FPN}}
    	\label{fpn}
    	\vspace{-.10in}
    \end{figure}
In figure \ref{fpn}, the bottom up and top-down architecture of  FPN is shown. The bottom-up pathway is the backbone CNN as used in Fast R-CNN. The top down path way consists a multilevel feature pyramid. Each level of top-down feature pyramid takes the upper layer's feature maps and the feature maps from corresponding stage of bottom-up pathway as input. Then concatenate those feature maps to produce the output. FPN also used RPN to generate region proposals. On the region proposal FPN is used for feature extraction.  RoIs are predicted  from different level to fuse the property of multi-scale and then those RoIs are reshaped using  RoI pooling layer as Fast R-CNN.  The output of RoI pooling layer is fed into FC layer for classification and bounding box regression.
     
\paragraph{\textbf{Training details:}}
The authors used ResNet based Fast R-CNN as the backbone network of FPN. Their network was end-to-end trainable using back propagation. They have used synchronized SGD with mini-batch size of 2 images and 256 anchors per image. For first 30K mini-batches the learning rate was 0.02 and 0.002 for next 10K mini-batches. The value of momentum and weight decay are 0.9 and 0.0001. The authors have trained their model using MS COCO dataset \cite{Lin2014} with 80 categories. 

%<<==========================Mask R-CNN==============================>>
\subsubsection{Mask R-CNN:}
He et al  extends previous  R-CNN object detection techniques to go one step further and locate exact pixels of each object instance (instance segmentation \cite{Iglovikov2018}\cite{Chen2017}) instead of just bounding boxes. As this model masks each instances of an object independently, they named it Mask R-CNN \cite{He2017}.  Mask R-CNN has the exact region proposal network(RPN) from Faster R-CNN to produce region proposals. The authors applied $RoIAlign$ layer on the region proposals instead of RoI pooling layer to align the extracted features with the input location of an object. The aligned RoIs are then fed into the last section of the Mask R-CNN to generate three output: a class label, a bounding box offset and a binary object mask. For masking, a small fully convolutional neural network is used on each RoI.  The architecture of Mask R-CNN is as figure \ref{fmask_rcnn}.
\begin{figure}[htb]
			\vspace{-.15in}
	\centering
	\includegraphics[scale=0.2]{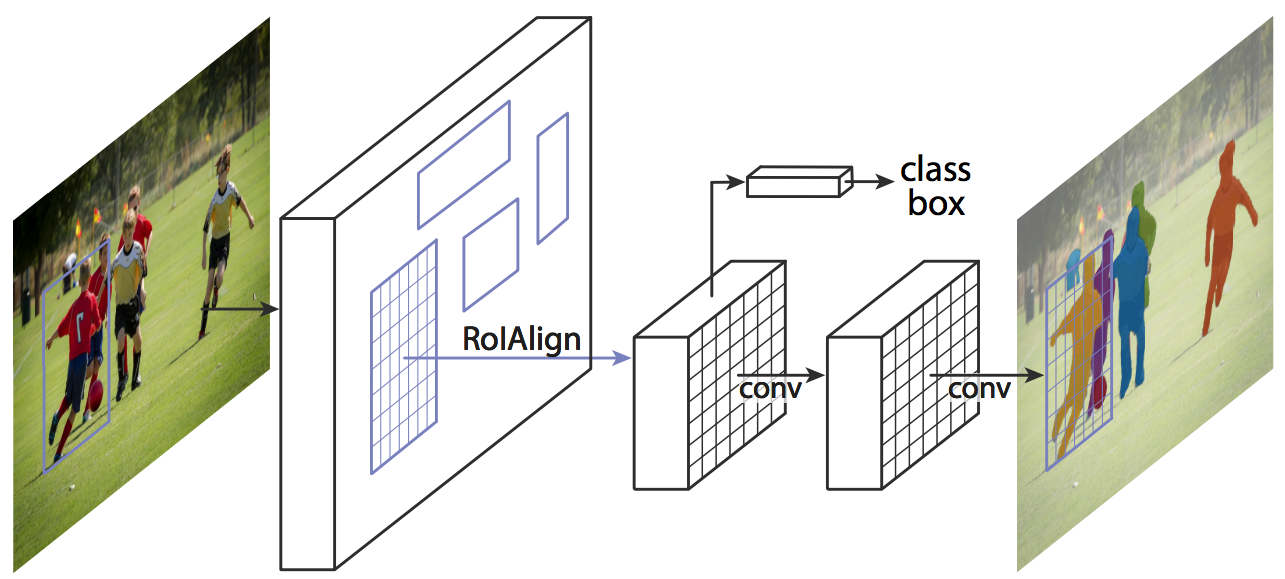}
	\caption{Mask R-CNN \cite{He2017}}
	\label{fmask_rcnn}
			\vspace{-.10in}
\end{figure}

\paragraph{\textbf{Training details:}}
The authors have used ResNet\cite{He2015} and ResNeXt\cite{Xie2016} networks as the base CNN of Faster R-CNN. Then Feature Pyramid Network(FPN) \cite{Lin2016_FPN} is used with Faster R-CNN as backbone. Region of interests(RoIs) are extracted form different levels of FPN with varying scales. Use of ResNet and FPN helped Mask R-CNN to gain better accuracy an speed. The authors set all hyperparameters of their model following \cite{Girshick2015}\cite{Ren2015}. If  the IoU of an RoI with a ground truth box is $\geq$ 0.5 then that RoI is considered as  positive otherwise negative. Mask R-CNN also followed image centric training. The multi-task loss function of Mask R-CNN combines the loss function of classification, localization and segmentation mask. The mask loss is defined only on positive RoIs. The ResNet\_FPN network is trained using SGD with a mini-batch size of 2 images. Each image is sampled into $N$ RoIs. The ratio of positive and negative RoIs for each mini-batch was 1:3. The learning rate for 160K iteration was 0.02. It was decreased by a factor of 10 for next 120K iteration. Also the momentum and weight decay was 0.9 and 0.0001 respectively. Training of ResNeXt\_FPN includes mini-batch size of 1 image and a initial learning rate of 0.01.

\subsection{One stage approach:}
In the one stage approach, classification and regression is done in a single shot using regular and dense sampling with respect to locations, scales and aspect ratio. The notable state-of-the-art models of one stage approach has covered here one-by-one: 
%<<==========================YOLO===================================>>
\subsubsection{YOLO:}
The You Only Look Once (YOLO) \cite{Redmon2015} is a single stage network model which predicts class probabilities and bounding boxes directly from input image using a simple CNN. The model divides the input image into fixed number of grids. Each cell of this grid predicts fixed number of bounding boxes with a confidence score. The confidence score is calculated by multiplying the probability to detect the object  with the IoU between the predicted and the ground truth boxes. The bounding boxes having the class probability above a threshold value is selected and used to locate the object within the image.
%\begin{figure*}
	%		\vspace{-.10in}
%	\subfloat[]{\includegraphics[width=5cm]{image/yolo}\label{fyolo}}
	%	\hspace{2mm}
%	\subfloat[]{\includegraphics[width=6cm]{image/ssd}\label{fssd}}\\
%	\caption{ Arcitecture of (a) YOLO \cite{Redmon2015} and (b)  }
	%	\label{fyolo_ssd}	
		%		\vspace{-.10in}
%\end{figure*}

\begin{figure}[htb]
	\vspace{-.15in}
	\centering
	\includegraphics[scale=0.6]{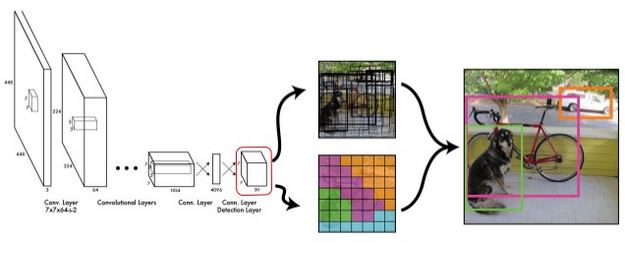}
	\caption{Arcitecture of  YOLO \cite{Redmon2015} }
	\label{fyolo}
	\vspace{-.10in}
\end{figure}
\paragraph{\textbf{Training details:}}
YOLO has 24 Conv layers and two FC layers. %For pre-training with ImageNet-2012 1000 class dataset, they have used 20 conv layers and added an average pooling layer and a FC layer. Then they have added 4 conv layers and two FC layers with randomly initialized weights. 
The authors have increased the resolution of input image and used batch normalization \cite{Bjorck2018} to the bounding box coordinates, box height and box width to increase detection accuracy. The final layer of the network used linear activation function and rest of the layers used leaky rectified linear activation \cite{Xu2015}. Uses of sum squared error in the output layer made their model to be optimized easily. Though YOLO predicts multiple bounding boxes per grid cell, the authors have chosen one object prediction per bounding box during training. These strategy leads to better prediction with respect to sizes, aspect ratio, classes and overall recall. They have trained/validated/tested their network on PASCAL VOC 2007 and 2012 datasets. For training, they have used mini-batch size of 64, momentum of 0.9 and weight decay of 0.005. Their initial learning rate was 0.001, then gradually increased it to 0.01 for 70 epochs, then decreased it by a factor of 10 for each 30 epochs. To avoid overfitting they have used dropout \cite{srivastava2014} and heavy data augmentation. 
%<<==========================SSD===================================>>
\subsubsection{SSD:}
The single shot multibox detection (SSD) \cite{Liu2016} model takes an entire image as input and passes it through multiple Conv layers with different sizes of filters ($10\times10$, $5\times5$ and $3\times3$) as shown in figure \ref{fssd}. Feature maps from conv layers at different position of the network are used to predict the bounding boxes. They are processed by a specific conv layers with $3\times3$ filters, called extra feature layers, to produce a set of bounding boxes. Similar to the anchor boxes (default box) of the Fast R-CNN, anchor box of SSD has parameters: the coordinates of the center, the width and the height. At the time of predicting bounding boxes, the model produces a vector of probabilities corresponding to the confidence over each class of object. In order to handle the scale, SSD predicts bounding boxes after multiple conv layers. Since each conv layer operates at a different scale, it is able to detect objects of various scales. This object detector achieves a good balance between speed and accuracy.
\begin{figure}[htb]
	\vspace{-.15in}
	\centering
	\includegraphics[scale=0.3]{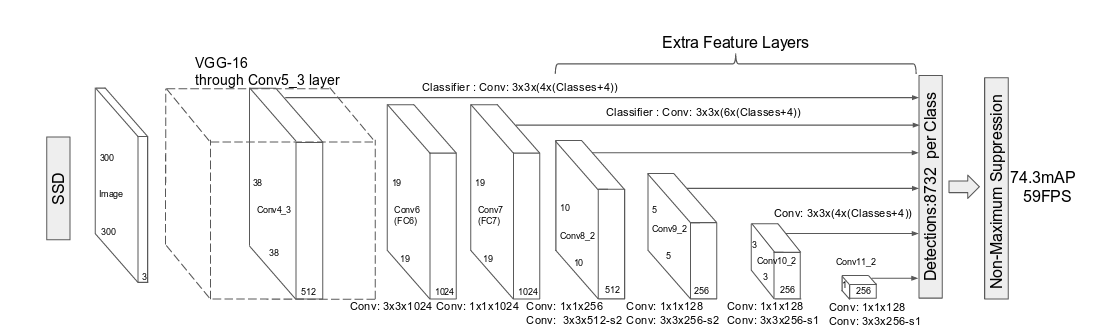}
	\caption{Arcitecture of  SSD \cite{Liu2016} }
	\label{fssd}
	\vspace{-.10in}
\end{figure}

\paragraph{\textbf{Training details:}}
During training, for each ground truth box, the authors have selected those anchor boxes that vary with respect to location, aspect ratio and scales. Then they have matched anchor with ground truth box with the best jaccard overlap \cite{Erhan2013}. They have selected those anchor boxes whose jaccard overlap to the ground truth is $>$ 0.5. To handle different object scales they have used different feature maps from different layers of their network. They have designed the anchors in such a way that certain feature maps learn to be responsive to particular scales of the objects. They have trained their model using hard negative mining with negative and positive default box ratio of 3:1 and using heavy data augmentation.
%<<==========================YOLO9000=================================>>
\subsubsection{YOLO9000:}
YOLO9000 \cite{Redmon2017} is a real-time object detector capable of detecting more than 9000 object categories by jointly optimizing detection and classification. The authors have used WordTree \cite{Wattenberg2008} hierarchy to combine data from various sources such as ImageNet and MS COCO \cite{Lin2014} dataset and proposed an algorithm to jointly train the network for optimization on those different dataset simultaneously. The authors also provide improvements over the initial YOLO model to enhance its performances without decreasing its speed.

\paragraph{\textbf{Training details:}}
The authors first produced YOLOv2 \cite{Nakahara2018} to improving the basic YOLO system. Then they have trained their model using joint training algorithm on more than 9000 classes from ImageNet and from MS COCO detection data. They have used batch normalization instead of dropout and their input image size was $416\times416$ to capture center part of an image. They have removed a pooling layer from the network to capture high resolution features. Anchor boxes are used for predicting bounding boxes. The dimensions of anchor box are chosen using K-means clustering on the training set. They have used fine grained features to detect smaller object and also used multi-scale training by choosing new image size randomly per 10 batches to make the model scale invariant.
%<<==========================RetinaNet=================================>>

\subsubsection{RetinaNet:}
RetinaNet \cite{Lin2017} is a simple one-stage, unified object detector which works on dense sampling of object locations in an input image. As shown in the figure \ref{fretinanet}, the model consists of a backbone network and two task-specific sub-networks. As backbone network, they have used feature pyramid network (FPN) \cite{Lin2016}. On the backbones output, the first sub-network performs convolutional object classification and the second sub-network performs convolutional bounding box regression. These sub-networks are basically small fully convolutional network (FCN) \cite{Shelhamer2017}  attached to each FPN level for parameter sharing. 
%\vspace{(-2cm)}
%\begin{figure*}
%			\vspace{-.10in}
%	\centering
%	\subfloat[]{\includegraphics[width=5cm]{image/retinanet}\label{fretinanet}}	\hspace{2mm}
%	\subfloat[]{\includegraphics[scale=0.2]{image/refineDet}\label{frefinedet}}\\
%	\caption{ Arcitecture of (a) RetinaNet \cite{Lin2017} and (b) RefineDet \cite{Zhang2017}}
	%	\label{fretinanet_refinedet}	
		%		\vspace{-.25in}
%\end{figure*}
\begin{figure}[htb]
	\vspace{-.15in}
	\centering
	\includegraphics[scale=0.3]{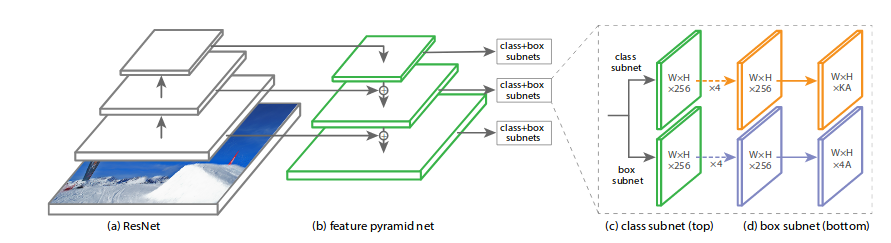}
	\caption{Arcitecture of  RetinaNet \cite{Lin2017} }
	\label{fretinanet}
	\vspace{-.10in}
\end{figure}

\paragraph{\textbf{Training details:}}
The authors have proposed a loss function called Focal loss to handle one stage object detection scenario where the network suffers from extreme foreground and background class imbalance during training. They have used this loss function on the output of the classification subnet. RetinaNet is trained with synchronized SGD over 8 GPUs with mini-batch of size 16 (2 images per mini-batch per GPU). The model is trained for 90K iterations with an initial learning rate of 0.01, which is then decreased by $1/10$ at 60K and again at 80K iterations. A weight decay of 0.0001 and a momentum of 0.9 are used. The training loss of their model is the sum of the focal loss and the standard smooth L1 loss used for box regression. 

%<<=========================RefineDet=================================>>
\subsubsection{RefineDet:}
RefineDet \cite{Zhang2017}, as shown in figure \ref{frefinedet} is a single-shot object detector based on a feed forward convolutional network. This network consists of two interconnected modules: $(i)$ the anchor refinement module (ARM) $(ii)$ the object detection module (ODM). The ARM module reduces search space for the classifier by filtering negative anchors and roughly adjusts the sizes and locations of the anchors for providing better initialization for the successive regressors. Taking the refined anchors from the previous module as input the ODM module improves the regression accuracy and predict multiple class labels. The transfer connected block (TCB) transfers the features in the ARM module to predict locations, sizes and class labels of object in the ODM module.
\begin{figure}[htb]
	\vspace{-.15in}
	\centering
	\includegraphics[scale=0.5]{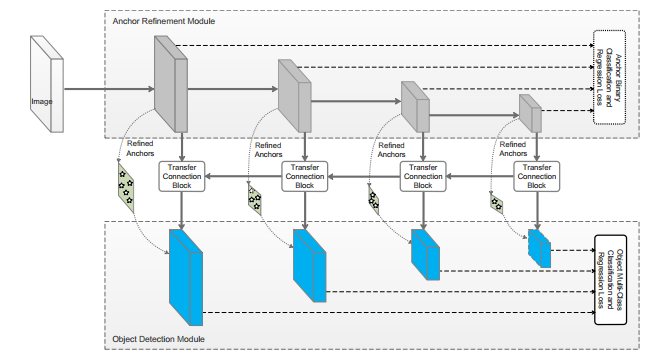}
	\caption{Arcitecture of RefineDet \cite{Zhang2017} }
	\label{frefinedet}
	\vspace{-.10in}
\end{figure}

\paragraph{\textbf{Training details:}}
The authors have used VGG-16 and ResNet 101 \cite{he16} network as base network. These networks are pre-trained on ImageNet classification and detection dataset. Instead of sixth and seventh FC layers of VGG-16, they have used Conv layers. Also to capture high level information and detecting objects in multiple scales, they have added two Conv layers at the end of the VGG-16 network and one residual block at the end of the ResNet-101. The parameters of extra two layers of VGG-16  are initialized using ``xavier" \cite{Glorot2010} method. To deal with different object scales, they have selected 4 feature layers with stride size 8, 16, 32 and 64 pixels for both CNN networks. Then they have associated those strides with several different anchors for prediction. The authors determined the matching of anchor and ground truth box using jaccard overlap following \cite{Liu2016}. Also they have trained their model using hard negative mining following \cite{Liu2016}.  They set the default training mini-batch size to 32 and then fine-tuned the network using SGD with 0.9 momentum and 0.0005 weight decay.

\section {Comparative Results}
As we have reviewed object detection models of two different types (two stage and one stage), performance comparison of those models is difficult task. In real life application, performance basically refers to accuracy and speed. We have seen here that  performance of a object detection model depends on different aspects such as feature extractor network, input image resolution, matching strategy and IoU threshold, Non-maximum suppression, the number of proposals or predictions, boundary box encoding, loss function, optimization function, chosen value for training hyperparameters  etc. In table \ref{table1}, we have shown comparative performance of different object detection models on PASCAL VOC and MS COCO  datasets. From second to sixth columns of the table represents the mean average precision (mAP) in measuring accuracy of object detection. Last two columns are giving a overview of those models speed and organizational approach. 
\begin{table*}

		\caption{Comparison of different object detection models: Column1 shows the name of the object detection model, Column 2 - 6 shows object detection accuracy (mAP) on different data set.  }
	\label{table1} 
%	\tiny
	\begin{tabular}{|c|c|c|c|c|c|c|c| }
		%\resizebox{0.8\textwidth}{!}{
		\hline
		\textbf{Object } & \textbf{PASCAL} & \textbf{PASCAL} & \textbf{PASCAL} & \textbf{COCO}  & \textbf{COCO}     & \textbf{Real}   & \textbf{Number}\\ 
		
		\textbf{Detection} & \textbf{VOC}  & \textbf{VOC} & \textbf{VOC} & \textbf{2015} & \textbf{2016}  & \textbf{time }& \textbf{of }\\
		
		\textbf{ Model}	& \textbf{2007} & \textbf{2010} & \textbf{2012} & \textbf{} & \textbf{}	&  \textbf{  Detection } & \textbf{Stages}\\
		
		\hline  \hline
		R-CNN	& 58.5\% & 53.7\% &53.3\%  &  &   & No &Two\\
		\hline
		
		SPP-net&59.2\%  &  &  &  &    &No  &Two\\
		\hline
		
		Fast&70.0\% &68.8\%  & 68.4\% &  &    &No &Two\\
		R-CNN&  &  &  &  &    &  &\\
		\hline

		Faster&73.2\%  &  &70.4\%  &  &  &No  &Two\\
		R-CNN&  &  &  &  &   &  &\\
		\hline
		 FPN& & & &  &    35.8\%&No &Two\\ \hline
		 
		Mask&  &  &  &  &  43.4\% &No &Two\\
	    R-CNN&  &  &  &  &    &  &\\	
		\hline
		
		YOLO&63.4\% &  &57.9\%  &  &    & Yes  &One\\
		\hline
		
		SSD& 81.6\%&  & 80.0\% &46.5\%  &   & No &One\\
		\hline
		YOLO9000& 78.6\%&  &73.4\%  &21.6\%  &  & Yes &One\\
		\hline
		RetinaNet&  &  &  &  &40.8\%  &Yes  &One\\
		\hline
		RefineDet& 85.6\% &  &86.8\% &  &   41.8\% & Yes  &One\\
		\hline
	\end{tabular}
	
\end{table*}

\section{Conclusion}
In this chapter, the advancements of different object detection models based on CNN has discussed. It has also shown that those models can be categorized into two different approaches: two-stage and one-stage. Two-stage models gave higher accuracy than one-stage models in object detection but they are slower. R-CNN, SPP-net and Fast R-CNN were slow because of external region proposal Network. Faster R-CNN overcome that problem using RPN. Mask R-CNN adds instance segmentation to the architecture of previous model. YOLO, SSD gave us a way for fast and robust object detection. RetinaNet focuses on improving loss function for better detection. RefineDet combined the merit of both two-stage and one-stage approach and it has achieved state-of the art performance. Through this chapter, the authors has also mentioned that the progress of various models are mainly because of better CNN models, new detection architecture, different pooling method, novel loss function etc. The improvements of different models give us hope towards more accurate and faster real time object detection.

%\begin{equation}
%\vspace{-.2in}
%g(x)=x^4+30x^3+216x^2+231x+116
%\end{equation}

\bibliographystyle{IEEEtran}

\bibliography{objdbib}

\end{document}